# Recombination vs Stochasticity: A Comparative Study on the Maximum Clique Problem

Michael Vella[1]
Department of CIS, Faculty of ICT, University of Malta, Msida, Malta

John Abela[1]
Department of CIS, Faculty of ICT, University of Malta, Msida, Malta
john.abela@um.edu.mt (corresponding author)

Kristian Guillaumier
Department of AI, Faculty of ICT, University of Malta, Msida, Malta

The maximum clique problem (MCP) is a fundamental problem in graph theory and in computational complexity. Given a graph $G$, the problem is that of finding the largest clique (complete subgraph) in $G$. The MCP has many important applications in different domains and has been much studied. The problem has been shown to be NP-Hard and the corresponding decision problem to be NP-Complete. All exact (optimal) algorithms discovered so far run in exponential time. Various meta-heuristics have been used to approximate the MCP. These include genetic and memetic algorithms, ant colony optimization, greedy algorithms, Tabu algorithms, and simulated annealing. This study presents a critical examination of the effectiveness of applying genetic algorithms (GAs) to the MCP compared to a purely stochastic approach. Our results indicate that Monte Carlo algorithms, which employ random searches to generate and then refine sub-graphs into cliques, often surpass genetic algorithms in both speed and capability, particularly in less dense graphs. This observation challenges the conventional reliance on genetic algorithms, suggesting a reevaluation of crossover and mutation operators' roles in exploring the solution space. We observe that, in some of the denser graphs, the recombination strategy of genetic algorithms shows unexpected efficacy, hinting at the untapped potential of genetic methods under specific conditions. This work not only questions established paradigms but also opens avenues for exploring algorithmic efficiency in solving the MCP and other NP-Hard problems, inviting further research into the conditions that favor purely stochastic methods over genetic recombination and vice versa.

## Background

Genetic Algorithms (GAs) and stochastic methods, such as Monte Carlo algorithms, are often used to solve (or approximate) NP-Hard optimization problems. GAs, inspired by Darwinian evolution and formalized by John Holland [Holland 1975], are adaptive search methods that use a population-based approach with genetic operators such as crossover and mutation. Stochastic methods, on the other hand, rely purely on random sampling to explore the solution space. In graph theory, the Maximum Clique Problem (MCP) is the problem of finding the largest complete subgraph within a given graph. This is known to be NP-hard. The DIMACS MCP 1 datasets provide standard benchmarks for evaluating these algorithms, facilitating comparative studies on their performance and efficacy.

## Genetic Algorithms

Genetic Algorithms (GAs) were inspired by the principles of natural selection and evolution as proposed by Charles Darwin. The idea is to mimic the process of natural selection where the fittest individuals are selected for reproduction to produce offspring of the next generation. This concept was first formalized in the 1960s by John Holland, a pioneer in the field of genetic algorithms. John Holland is credited with the foundational work on genetic algorithms. His seminal book *Adaptation in Natural and Artificial Systems* [Holland 1975] laid the groundwork for the development of GAs. Holland's work focused on the mechanisms of adaptation in both natural and artificial systems, exploring how the principles of natural evolution can be applied to solve, or approximate, complex problems algorithmically.

---

[1] Joint first authors.

### Initial Population

The process begins with a randomly generated population of potential solutions, referred to as chromosomes or individuals. Each chromosome in the population is a candidate solution to the problem. The chromosome representation can vary but typically involves encoding the solution as a string (or vector) of genes. For example, in binary encodings, a chromosome consists of a string (or vector) of 0s and 1s.

### Fitness Function

The fitness function evaluates each chromosome's performance or 'fitness' in solving the problem. This function is crucial as it directly influences the selection of individuals for reproduction and survival. The fitness function must accurately reflect the quality of the solutions. A poorly designed fitness function can mislead the algorithm, leading to sub-optimal solutions or premature convergence.

### Genetic Operators

Genetic operators are used in GAs to explore and exploit the search space efficiently and effectively.

- **Crossover (Recombination)** Combines parts of two parent solutions to generate offspring, promoting diversity and potentially combining beneficial traits from both parents. The idea is to exchange segments of parents' chromosomes to create new offspring with potentially better performance. Different crossover methods, such as single-point or multi-point crossover, can be used.

- **Mutation** Introduces random changes to individual solutions, maintaining genetic diversity within the population and preventing premature convergence to sub-optimal solutions.

These operators must be tailored to suit the problem's structure. For instance, in the Maximum Clique Problem, simply combining two cliques through crossover might not yield a larger clique, especially in sparser graphs. Hence, careful design of crossover mechanisms is crucial. Also, the mutation rate needs to be balanced to maintain diversity without disrupting good solutions. The wrong choice of genetic operators may lead to a slow, ineffective search of the solution space.

If the genetic operators and fitness function are not optimally chosen, the GA may fail to find good solutions, converge prematurely, or require excessive computational resources. This can result in inefficiency and ineffectiveness, demonstrating the critical importance of these choices in the success of GAs.

### Selection and Evolution

The population is expanded by adding offspring using the genetic operators. Selection is then used to reduce the size of the expanded population. Selection mechanisms choose individuals from the current population to be parents for the next generation. Methods such as roulette wheel selection, tournament selection, or rank-based selection can be used. The process of selection, crossover, and mutation is repeated over many generations. Each new generation is expected to be better than the previous one, gradually evolving towards an optimal or near-optimal solution.

## Monte Carlo Algorithms

Stochastic methods [Marti 2024] are computational techniques that, in contrast to purely deterministic approaches, employ randomness to solve or approximate complex problems. These techniques are particularly valuable for exploring high-dimensional spaces and avoiding local optima, making them robust in diverse applications.

Monte Carlo algorithms [Fishman 2011], a subset of stochastic methods, use random sampling to estimate numerical results for optimization problems. Named after the Monte Carlo Casino in Monaco, these algorithms are popular due to their intrinsic simplicity and effectiveness in optimization problems and are particularly effective for problems involving complex, high-dimensional spaces where other meta-heuristics may not perform well. The inherent randomness in stochastic methods allows for a broader exploration of the solution space, potentially avoiding local optima and providing more robust solutions. In the simplest form, Monte Carlo algorithms generate random solutions independently. Each solution is evaluated using a predefined criterion, but no selection or adaptation process occurs based on these evaluations. This approach is purely stochastic, relying on the law of large numbers to achieve accurate results over numerous iterations.

## The Maximum Clique Problem (MCP)

### Preliminary Definitions

1. A **graph** $G = (V, E)$ consists of a non-empty set $V$ of vertices (or nodes) and a set $E$ of edges. Each edge $e \in E$ is a two-element subset (pair) of $V$, i.e., $e = (u, v)$ where $u, v \in E$. The graph $G$ represents a collection of vertices connected by edges. Note that $E \subseteq V \times V$.

1. Let $G = (V, E)$ be a graph. A graph $H = (V', E')$ is a **subgraph** of $G$ if $V' \subseteq V$ and $E' \subseteq E$. In other words, $H$ is obtained from $G$ by selecting a subset of vertices and edges from $G$.

2. A graph, or subgraph, is **complete** if every pair of vertices is pairwise adjacent. In other words, every pair of vertices is connected by an edge.

3. Let $G = (V, E)$ be an undirected graph. A **clique** $C$ in $G$ is a subset $C \subseteq V$ of vertices of $G$, such that for every pair of distinct vertices $u, v \in C$, there exists an edge $e = (u, v) \in E$. In other words, every vertex in $C$ is connected to every other vertex in $C$. A subgraph of $G$ of size $n$ is a clique if all the vertices have a degree of $n - 1$.

4. A **maximal clique** $C$ in a graph $G$ is a clique that cannot be extended by adding any vertex from $G$ without violating the condition of being a clique. In other words, if $C$ is a clique and there exists a vertex $v \in V$ such that $C \cup v$ is also a clique, then $C$ is not a maximal clique.

5. A **maximum clique** $C$ in a graph $G$ is a clique that contains the largest number of vertices among all cliques in $G$. In other words, it is the clique of maximum cardinality.

Note that every maximum clique is maximal but not every maximal clique is necessarily maximum.

### Finding the Maximum Clique in a Graph

The maximum clique problem (MCP) is that of finding the largest clique in a given graph $G$. The MCP has been extensively studied and has numerous applications in science, engineering, and industry. These applications include telecommunications, bioinformatics, social networks, graph colouring, information retrieval, computer vision, cluster analysis, optimization problems, transportation and logistics, and information security. The MCP was shown to be NP-Hard and the corresponding decision problem shown to be NP-Complete by Karp [Karp 1972]. The MCP can also easily be shown to be NP-Hard by Turing reduction from either the *vertex cover* (VC) or the *independent set* (IS) problems. A consequence of this is that all known exact (optimal) algorithms run in exponential time. There cannot be a polynomial-time optimal algorithm for MCP unless $P = NP$. MCP is not only very hard to solve optimally but it is also hard to approximate. In fact, no approximation algorithm in the literature has a constant approximation ratio. An exposition and a proof of the non-approximability of MCP can be found in [Ausiello et al. 1999].

Various algorithms have been proposed to find all the maximal cliques in a given undirected, simple, graph. One of the most well-known is the Bron-Kerbosch algorithm [Bron and Kerbosch 1973]. This is a recursive back-tracking algorithm that finds all maximal cliques in a given graph $G$. This is a brute-force algorithm, and the time complexity can be exponential, $O(3^{n/3})$, in the worst case.

Various meta-heuristics have been applied to the problem of approximating the MCP. The reader is referred to [Wu and Hao 2015; Singh 2015; Fakhfakh et al. 2018] for expositions. Apart from genetic and memetic algorithms, these include ant colony optimization (ACO) [Soleimani-Pouri, Rezvanian, and Meybodi 2012], Branch and Bound algorithms [Morrison et al. 2016], local search techniques [Katayama, Hamamoto, and Narihisa 2005; Battiti and Protasi 2001], and artificial neural networks [Yang et al. 2009].

### MCP Datasets

The DIMACS (Center for Discrete Mathematics and Theoretical Computer Science) Network Data Repository is a graph and network repository that contains hundreds of real-world networks and benchmark graph datasets [Rossi and Ahmed 2015]. The repository is curated by Rutgers University.

**Table 1**. Properties of the subset of graphs from the DIMACS benchmark graphs used in [Zhang et al. 2014]

| Graph | Vertices | Edges | Density | LC Known |
|---|---:|---:|---:|---:|
| C125.9 | 125 | 6,963 | 0.898 | 34 |
| C250.9 | 250 | 27,984 | 0.899 | 44 |
| C500.9 | 500 | 112,332 | 0.900 | 57 |
| C1000.9 | 1,000 | 450,079 | 0.901 | 68 |
| C2000.9 | 2,000 | 1,799,532 | 0.900 | 80 |
| C2000.5 | 2,000 | 999,836 | 0.500 | 16 |
| C4000.5 | 4,000 | 4,000,268 | 0.500 | 18 |
| DSJC500 | 500 | 125,248 | 0.502 | 13 |
| DSJC1000 | 1,000 | 499,652 | 0.500 | 15 |
| brock200 | 200 | 9,876 | 0.496 | 12 |
| brock200 | 200 | 13,089 | 0.658 | 17 |
| brock400 | 400 | 59,786 | 0.749 | 29 |
| brock400 | 400 | 59,765 | 0.749 | 33 |
| **brock800** | 800 | 208,166 | 0.651 | 24 |
| **brock800** | 800 | 207,643 | 0.650 | 26 |
| **gen200p0.9** | 200 | 17,910 | 0.900 | 44 |
| **gen200p0.9** | 200 | 17,910 | 0.900 | 55 |
| **gen400p0.9** | 400 | 71,820 | 0.900 | 55 |
| **gen400p0.9** | 400 | 71,820 | 0.900 | 65 |
| **gen400p0.9** | 400 | 71,820 | 0.900 | 75 |
| **hamming8-4** | 256 | 20,864 | 0.639 | 16 |
| **hamming10-4** | 1,024 | 434,176 | 0.829 | 40 |
| **keller4** | 171 | 9,435 | 0.649 | 11 |
| **keller5** | 776 | 225,990 | 0.752 | 27 |
| **keller6** | 3,361 | 4,619,898 | 0.818 | 59 |
| **p-hat300-1** | 300 | 10,933 | 0.244 | 8 |
| **p-hat300-2** | 300 | 21,928 | 0.489 | 25 |
| **p-hat300-3** | 300 | 33,390 | 0.744 | 36 |
| **p-hat700-1** | 700 | 60,999 | 0.249 | 11 |
| **p-hat700-2** | 700 | 121,728 | 0.498 | 44 |
| **p-hat700-3** | 700 | 183,010 | 0.748 | 62 |
| **p-hat1500-1** | 1,500 | 284,923 | 0.253 | 12 |
| **p-hat1500-2** | 1,500 | 568,960 | 0.506 | 65 |
| **p-hat1500-3** | 1,500 | 847,244 | 0.754 | 94 |

The DIMACS dataset is widely used by researchers and allows direct comparison of the performance and results of different meta-heuristics that approximate the MCP and other NP-Hard graph-theoretic problems. This dataset includes several different classes of graphs ranging from random graphs with a known maximum clique, to graphs obtained from various domains. Table 1 shows the properties and attributes of these benchmark graphs. The number of vertices and edges are displayed alongside the density (rounded to 3 decimal places) and the best result known of the maximum clique of that graph instance.

The table only shows the subset of the DIMACS dataset used by Zhang *et al* and other researchers [Zhang et al. 2014; Kizilates-Evin 2020; Moussa, Akiki, and Harmanani 2019]. Some of the higher cardinality graphs were not included. These were probably excluded for computational reasons.

## Survey of Existing Literature

One of the first papers to describe the application of genetic algorithms to find cliques in graph was that of [Murthy, Parthasarathy, and Sastry 1994]. Just a year later, the authors of [Bui and Eppley1995] proposed a hybrid genetic algorithm, which they called GMCA, to approximate the MCP. The GMCA algorithm incorporated a local search optimization technique which ran at each generation and included a pre-processing stage which ordered the vertices by degree, making higher order vertices closer to each other with the aim of making crossover less likely to disrupt potential solutions. The GMCA algorithm was tested on some of the graphs in the DIMACS dataset with generally positive results, although certain graph characteristics led to poorer performance. In 1998, [E. Marchiori 1998] proposed a novel heuristic genetic algorithm called HGA. Uniform crossover, swap mutation, and elitism selection are incorporated in Marchiori's algorithm. The novelty in this algorithm is in the repair and clique extension methods used. Chromosomes encode cliques and once the crossover and mutation genetic operator is applied to a chromosome the offspring generated are very often not cliques. Hence, the author developed a repair method in which vertices from a chromosome are removed in such a way that the chromosome is reduced to a clique. Afterwards, a chromosome extension method is used to extend the repaired chromosome by adding more vertices to it to generate a maximal clique. Satisfactory results were obtained when evaluating the HGA on the DIMACS benchmark graph instances and the HGA algorithm achieved better results than the GMCA algorithm [Bui and Eppley 1995].

In 2002, two genetic algorithms for approximating the MCP were proposed in [Huang 2002; Marchiori 2002]. The algorithm in [Huang 2002] is a variation of the GMCA [Bui and Eppley 1995] algorithm. Both algorithms use graph pre-processing and incorporate a local search technique. Furthermore, since the crossover and mutation genetic operators generate new offspring which are more likely to be non-cliques, the algorithm, which was called HGAMC, incorporates two procedures similar to the repair and chromosome extension methods used in HGA [E. Marchiori 1998] as well as the clique extraction and clique improvement methods in order to speed up convergence and generate maximal cliques. Moreover, the initial population is created by a greedy algorithm by picking a random vertex from the graph, iterating randomly over the adjacent vertices, and adding that specific vertex if it is connected to all the current nodes in the chromosome. This results in the initial population being made up of chromosomes that encode actual, but not necessarily maximal, cliques. The HGAMC algorithm was evaluated on the DIMACS benchmark instances and provided satisfactory results, some results being similar to what were achieved in the HGA [E. Marchiori 1998]. The GENE genetic algorithm proposed in [Marchiori 2002] is based on the previous HGA [E. Marchiori 1998] algorithm but the logic for the repair and extend operators is slightly different. Furthermore, the GENE algorithm uses different parameters (crossover rate, mutation rate, etc.) from the HGA. The GENE algorithm was compared to an iterated local search algorithm, ITER and a multi-start local search algorithm, MULTI on the DIMACS graphs. The ITER and GENE algorithms yielded good results, with ITER outperforming GENE in five instances and GENE outperforming ITER in four instances.

In 2005, an evolutionary algorithm with guided mutation, EA/G [Zhang, Sun, and Tsang 2005] was proposed. The EA/G algorithm also uses the repair method developed in HGA, and the guided mutation operator performs single bit flip mutation to the best chromosome in the population, as the resultant solutions are not far from the best solution found so far. The algorithm also divides the search space into several search areas, and then focuses on different areas at different times. The EA/G algorithm was also evaluated on the DIMACS graphs and provided the best solutions found with a genetic algorithm as of that time.

In 2007, a different type of genetic algorithm [Ouch, Reese, and Yampolskiy 2007] which combines two populations was proposed. The first population was based on a genetic algorithm while chromosomes from the second sub-population were derived using a local search technique. Then, at each generation the best chromosomes from the sub-population replace a percentage of the worst individuals from the main population. The algorithm also uses a different fitness technique, known as the fitness sharing method. The fitness sharing method helps to diversify the population by reducing the fitness of individuals in populated landscapes. Comparing results with the EA/G algorithm on a subset of the DIMACS instances, this algorithm obtained worse results.

Two genetic algorithms [Bhasin and Mahajan 2012; Bhasin, Kumar, and Munjal 2013] were published in 2012 and 2013. Both algorithms do not seem to use some form of a repair method to repair chromosomes after crossover or mutation. The authors state that, for graphs with more than 70 vertices, the algorithm obtains unsatisfactory results. Furthermore, these algorithms were not evaluated on any of the DIMACS graphs.

In 2014 two genetic algorithm approaches to approximating the MCP, [Savita 2014; Zhang et al. 2014] were proposed. The author in [Savita 2014] uses a strong local optimizer to accelerate the convergence and the adaptive genetic operators to further improve the algorithm, however, these changes are not explicitly explained in the paper. Furthermore, the algorithm provides no test results on the DIMACS benchmark graphs. Zhang et al. in [Zhang et al. 2014] created a fast genetic algorithm, FGA, that obtained better results than all its predecessors when evaluated in the DIMACS graphs. The genetic operators used were uniform crossover, inversion mutation, and random selection with elitism. Moreover, the algorithm uses a greedy repair and chromosome extension techniques by always deleting the node with the lowest degree for repairing and selecting that node with the highest degree when extending. In [Zhang et al. 2014] the authors also compare the time efficiency between their FGA and the GENE algorithms. It was reported that FGA performed substantially better.

In 2015, a new parallel hybrid genetic algorithm on OpenCL was proposed [Li et al. 2015]. The algorithm uses multi-point crossover, inversion mutation and repair and extend techniques very similar to those in the HGA algorithm [E. Marchiori 1998]. This was the first time that the selection, the crossover and mutation operators, and the repair and extraction methods were parallelized.

Two new genetic algorithm approaches to approximating the MCP were proposed in 2019 [Moussa, Akiki, and Harmanani 2019; Fallah, Keshvari, and Fazlali 2019]. The algorithm in [Fallah, Keshvari, and Fazlali 2019] is similar to the algorithm proposed in [Zhang et al. 2014] but parallelized. Instead of random selection with elitism, roulette wheel selection with elitism is used. Furthermore, three different repair methods are defined and chosen randomly at each step. The algorithm proposed by Moussa et al. in [Moussa, Akiki, and Harmanani 2019] uses a new crossover genetic operator in which a bitwise AND is applied between both parents, thus only the common vertices remain in the offspring. Other genetic operators used are roulette wheel selection and one-bit flip mutation. An *extend-clique* method is also used to extend the offspring (after crossover) and turn them into maximal cliques. This algorithm was evaluated on the DIMACS benchmarks graphs, obtaining satisfactory results, very similar to what were obtained in the FGA algorithm [Zhang et al. 2014].

In 2020, three new papers were published. The authors in [Sahani et al. 2020] developed a genetic algorithm for finding all maximal cliques in a graph. The algorithm in [Shah 2020] is an open-source genetic algorithm using the same bitwise AND crossover as defined in [Moussa, Akiki, and Harmanani 2019]. The authors do not provide any information regarding the other genetic operators used apart from the application of mutation, where mutation is only performed if the fitness of the offspring is less than or equal to at least one of the parent solutions. The authors in [Kizilates-Evin 2020] propose a new genetic algorithm, rHGA, which uses a new type of genetic repair method in which chromosomes are repaired, extended, and vertices are sometimes then swapped. Satisfactory results were obtained on the DIMACS benchmark graphs, comparable to those obtained in [Zhang et al. 2014; Moussa, Akiki, and Harmanani 2019].

## The Zhang et al Paper of 2014

The paper [Zhang et al. 2014] was chosen as the preferred baseline for comparison to stochastic Monte Carlo methods. The main reason for this is because this paper gives a detailed description of the genetic algorithm used as well as its parameters, including the genetic operators, chromosomal representation, and the fitness function. This made it straightforward to reproduce the Zhang FGA algorithm. The original code used by Zhang was not made available. The Zhang FGA algorithm was implemented in Python 3.11 based on the description of the algorithm and its various parameters in the [Zhang et al. 2014] paper. Given a graph $G$, the Zhang GA attempts to find the largest possible clique in reasonable time. As with all genetic algorithms, the larger the time budget, the longer the algorithm is run, and this increases the probability of finding a solution that is closer to the optimal for the problem at hand.

### Chromosomal Representation

In the Zhang implementation, a chromosome represents a subgraph of the input graph $G$. In essence, a chromosome is just a subset of the vertices encoded as a binary vector [Zhang et al. 2014]. The dimension of the vector is equal to the order of the graph and a gene can have either a value of 0 or 1 where 1 denotes that the corresponding vertex is in the subgraph and 0 otherwise. An example of a binary vector chromosomal representation is shown in Figure 1 below.

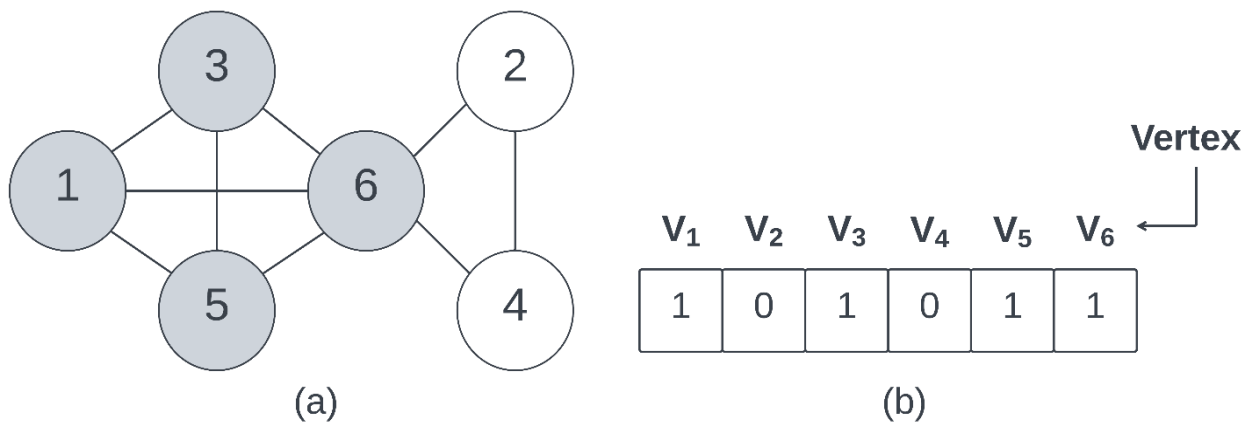

**Figure 1** (**a**) Example graph $G$, (**b**) Chromosome binary vector encoding for the subgraph of $G$ (highlighted in grey).

### Chromosome Generation

A chromosome is generated by sampling a uniform distribution to create a binary vector. Specifically, each bit in the vector is independently sampled from a Bernoulli distribution with parameter $p = 0.5$, ensuring an equal probability of being 0 or 1. This binary vector encodes a random subgraph of $G$, $G_S$. Note that, in general, $G_S$ may not be a clique.

### Chromosome Repair

A chromosome that is either randomly generated as described above, or formed as a result of mutation or crossover, might not encode a clique. The repair procedure, described below, is based on that in [E. Marchiori 1998] and is then used to repair the chromosome to restore the clique property.

1. $G_S$ is therefore 'reduced' to a clique using a greedy procedure by iteratively removing the vertex with the lowest degree until the resulting chromosome encodes a clique.

1. $G_S$ is then extended to a maximal clique by iteratively adding vertices from $G - G_S$ that are adjacent to all the vertices in $G_S$. When no such vertices remain in $G - G_S$, then $G\_S$ is a maximal clique.

### Fitness Function

The fitness function used in [Zhang et al. 2014] is very simple. It is essentially the cardinality of the clique encoded in the chromosome.

$$f(X) = \begin{cases} \sum_{i=1}^{n} x_i \ \text{if chromosome is a clique}, \\ \quad\quad 0 \ \text{otherwise}. \end{cases}$$

The fitness of a chromosome is therefore equal to the size (in vertices) of the subgraph it represents (i.e., the number of 1's occurring in the vector string) if the subgraph is a clique. Else, if the subgraph does not form a clique, the fitness of the chromosome is 0.

### Genetic Operators

The Zhang paper uses the uniform crossover proposed in [Syswerda 1989]. A mask chromosome is created. The properties of the mask chromosome are that it should be equal to the size of the parent chromosomes and contains only the values 0 or 1, randomly generated at each gene. At each position where the mask contains the value of 1, the parents swap the value of their genes at that position. Each pair of parents produces 2 offspring to form a family.

Zhang *et al* used a simple inversion mutation operator where two points are randomly selected from the chromosome and all genes between these two points are inverted, meaning that the first bit switches places with the last bit, the second bit switches places with the second to last bit, repeating this process until all bits have been inverted.

### Elitist Selection

A generation is expanded by shuffling the population and creating mating pairs of chromosomes. Uniform crossover is used to create two offspring from each mating pair to form a family. The inverse mutation operator is applied to each offspring to diversify the chromosomes. Chromosome repair is finally applied to each offspring. The fittest two chromosomes for each family are chosen for the next generation.

### FGA Parameters

In the Zhang *et al* implementation, the generation size was set to 50 and the maximum number of generations was set to 100. The FGA was run 100 times for each of the 34 selected graphs in the DIMACS dataset.

## The Monte Carlo Algorithm

The Monte Carlo algorithm we created stochastically traverses the search space by randomly generating a population of chromosomes and then returns the largest clique found. Three different methods are used to generate the chromosomes.

1. Generate chromosomes randomly by flipping a (uniform) coin at each gene of the chromosome and then using the repair and extend methods to form maximal cliques. This is the method used in [Zhanget al. 2014].

1. Generate chromosomes by randomly selecting an edge of the graph and using the *clique-extend* method to (stochastically) extend that edge to a maximal clique.

2. Generate chromosomes by first randomly selecting a vertex from the input graph and then adding all the neighbours of that vertex. The resulting chromosome is then reduced to a clique and then extended to a maximal clique using *clique extend*.

For methods 1 and 3, after the chromosomes is repaired, the newly formed clique may be pruned by randomly deleting some vertices from that chromosome. This is to expand the search space since smaller cliques are more likely to be sub-cliques of the target clique.

## Recombination vs Stochasticity

The focus of this work was to investigate whether the genetic operators for recombination in genetic algorithms used in the literature actually perform better than random search. To this effect we reproduced, and parallelized, the genetic algorithm described in [Zhang et al. 2014] for the purpose of reproducing the results reported by Zhang. We also developed a Monte Carlo (MC) algorithm that searched for the largest cliques in the DIMACS dataset using purely random search. In both the Zhang FGA and the Monte Carlo algorithm, a chromosome is a binary vector that encodes the list of vertices that comprise a clique in the graph. In the genetic algorithm, the fittest chromosomes (those that represent the largest cliques) are used to create new chromosomes (offspring) by combining (through crossover) genes from both parents in the hope of producing fitter offspring while in the Monte Carlo algorithm there is no recombination (crossover) or mutation. Chromosomes are generated completely randomly.

### Experimentation Results
*The FGA vs MC Results*

We ran the replicated Zhang FGA and the MC algorithms three times on each of the DIMACS datasets. The seeds of the random number generator were changed for every run. The results are shown in Figure 2 below. For each algorithm, we picked the best results out of the three runs. We also show the average number of chromosomes considered by each method until the best result is found.

As shown in Figure 2, the Zhang FGA found a larger clique for three graphs. This was for the *C1000.9*, *C2000.9* and *gen400_p0.9_65* graphs (highlighted in blue). On the other hand, The MC algorithm found larger cliques in 6 graphs (highlighted in green). The cells highlighted in orange denote the cases when an algorithm considered (generated) less chromosomes to find the largest clique. The Zhang FGA algorithm found the largest maximal clique faster in 8 graphs while the MC algorithm returned the largest found clique faster in the remaining 26 DIMACS graphs. One must point out that, in two cases, for the graphs *gen400 p0.9 55* and *keller6*, the clique sizes found by Zhang *et al* were larger than we found using either the replicated Zhang FGA or the MC algorithm.

| Graph | Nodes | Edges | Density | FGA 3 Run Average | | Monte Carlo 3 Run Average | | Zhang | Best |
|---|---|---|---|---|---|---|---|---|---|
| | | | | Max | Chromosomes | Max | Chromosomes | | |
| C125.9 | 125 | 6963 | 0.898 | 34 | 113 | 34 | 46 | 34 | 34 |
| C250.9 | 250 | 27984 | 0.899 | 44 | 8,867 | 44 | 1,595 | 44 | 44 |
| C500.9 | 500 | 112332 | 0.9 | 57 | 147,458 | 57 | 36,020 | 56 | 57 |
| C1000.9 | 1000 | 450079 | 0.901 | 67 | 379,135 | 66 | 143,089 | 65 | 68 |
| C2000.9 | 2000 | 1799532 | 0.9 | 75 | 238,288 | 74 | 27,355 | 74 | 80 |
| DSJC500_5 | 500 | 125248 | 0.502 | 15 | 121,015 | 15 | 851 | 15 | 15 |
| DSJC1000_5 | 1000 | 499652 | 0.5 | 13 | 7,630 | 13 | 93 | 13 | 13 |
| C2000.5 | 2000 | 999836 | 0.5 | 16 | 74,217 | 16 | 3,502 | 16 | 16 |
| C4000.5 | 4000 | 4000268 | 0.5 | 17 | 110,605 | 17 | 6,125 | 17 | 18 |
| brock200_2 | 200 | 9876 | 0.496 | 12 | 174,045 | 12 | 575 | 12 | 12 |
| brock200_4 | 200 | 13089 | 0.658 | 16 | 1,668 | 17 | 6,510 | 16 | 17 |
| brock400_2 | 400 | 59786 | 0.749 | 25 | 81,173 | 29 | 11,579 | 25 | 29 |
| brock400_4 | 400 | 59765 | 0.749 | 25 | 23,050 | 33 | 2,171 | 25 | 33 |
| brock800_2 | 800 | 208166 | 0.651 | 21 | 125,993 | 21 | 9,114 | 21 | 24 |
| brock800_4 | 800 | 207643 | 0.65 | 21 | 266,016 | 21 | 10,965 | 20 | 26 |
| gen200_p0.9_44 | 200 | 17910 | 0.9 | 44 | 14,435 | 44 | 4,899 | 44 | 44 |
| gen200_p0.9_55 | 200 | 17910 | 0.9 | 55 | 738 | 55 | 189 | 55 | 55 |
| gen400_p0.9_55 | 400 | 71820 | 0.9 | 52 | 89,909 | 53 | 151,776 | 55 | 55 |
| gen400_p0.9_65 | 400 | 71820 | 0.9 | 65 | 21,141 | 64 | 29,631 | 65 | 65 |
| gen400_p0.9_75 | 400 | 71820 | 0.9 | 75 | 5,646 | 75 | 885 | 75 | 75 |
| hamming8-4 | 256 | 20864 | 0.639 | 40 | 14,080 | 40 | 861 | 40 | 40 |
| hamming10-4 | 1024 | 434176 | 0.829 | 16 | 3 | 16 | 1 | 16 | 16 |
| keller4 | 171 | 9435 | 0.649 | 11 | 124 | 11 | 2 | 11 | 11 |
| keller5 | 776 | 225990 | 0.752 | 27 | 28,081 | 27 | 64 | 27 | 27 |
| keller6 | 3361 | 4619898 | 0.818 | 55 | 232,600 | 56 | 15,958 | 57 | 59 |
| p_hat300-1 | 300 | 10933 | 0.244 | 8 | 164 | 8 | 140 | 8 | 8 |
| p_hat300-2 | 300 | 21928 | 0.489 | 25 | 109 | 25 | 82 | 25 | 25 |
| p_hat300-3 | 300 | 33390 | 0.744 | 36 | 602 | 36 | 2,201 | 36 | 36 |
| p_hat700-1 | 700 | 60999 | 0.249 | 11 | 710 | 11 | 5,236 | 11 | 11 |
| p_hat700-2 | 700 | 121728 | 0.498 | 44 | 258 | 44 | 2,468 | 44 | 44 |
| p_hat700-3 | 700 | 183010 | 0.748 | 62 | 2,589 | 62 | 492 | 62 | 62 |
| p_hat1500-1 | 1500 | 284923 | 0.253 | 11 | 784 | 12 | 1,778 | 11 | 12 |
| p_hat1500-2 | 1500 | 568960 | 0.506 | 65 | 576 | 65 | 25,330 | 65 | 65 |
| p_hat1500-3 | 1500 | 847244 | 0.754 | 94 | 11,399 | 94 | 8,707 | 94 | 94 |
| | | | | Avg | 64,212 | Avg | 15,009 | | |

**Figure 2** The FGA vs the MC Algorithm Results

The results also show that, in general, the MC algorithm generated much fewer (by a factor of 4) chromosomes before it returned its best result. The average density of the graphs where the FGA found a larger clique is 0.9 and the average density for the graphs where the MC algorithm found a larger clique is 0.69.

In order to determine why, in some cases, combination worked better than stochastic search we performed the following *post factum* analysis. We plotted the frequency of the nodes in the best chromosome (i.e. the highest cardinality clique) against the generations.

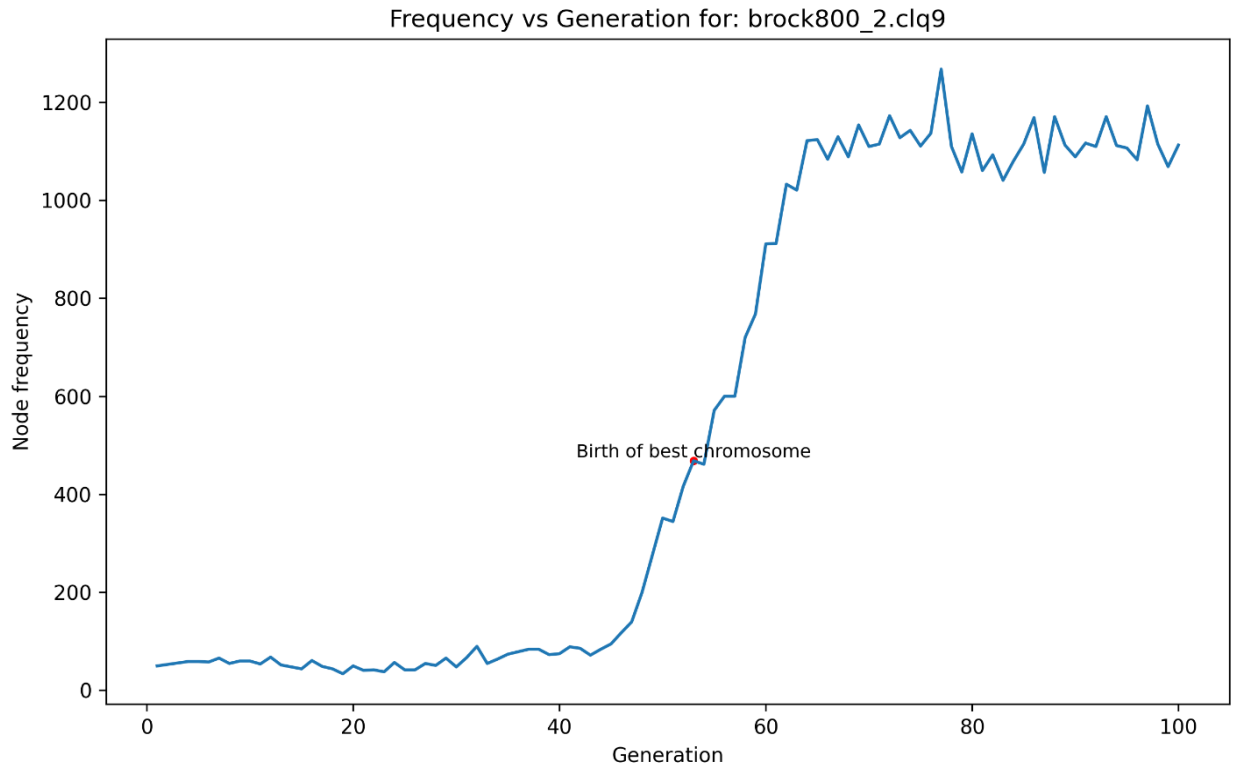

**Figure 3** Best chromosome node frequency in the GA population (all generations) for *brock800_2*

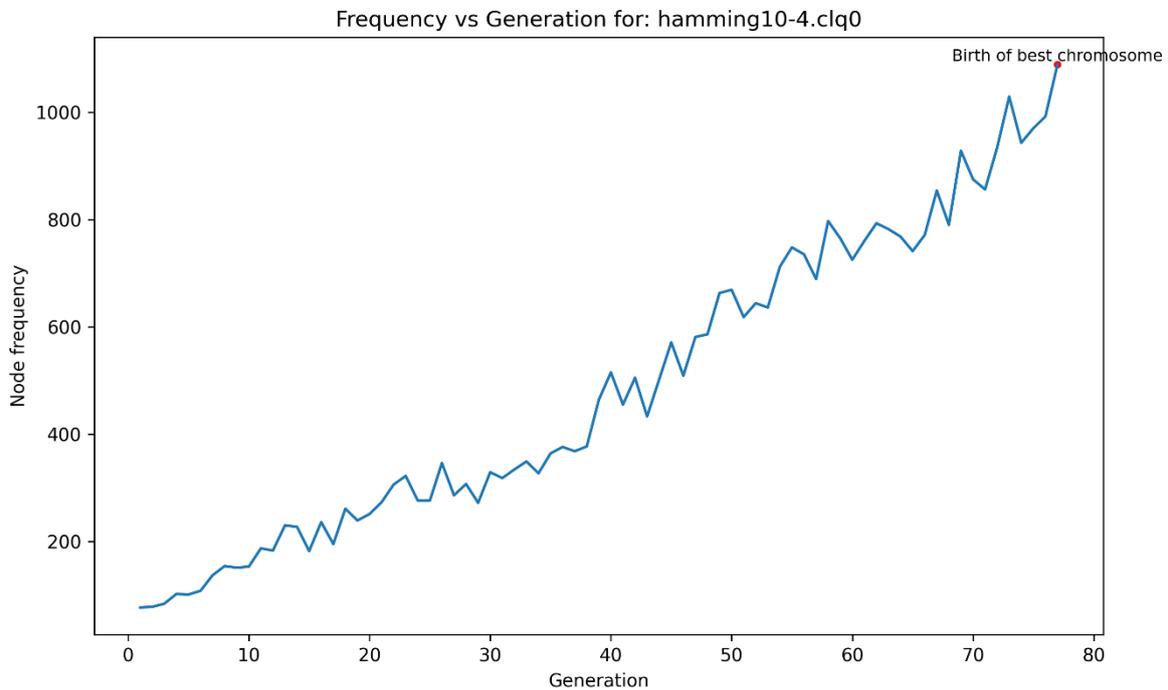

**Figure 4** Best chromosome node frequency in the GA population (all generations) for *hamming10-4*

In other words, for each generation, we determined the frequency of the nodes of the best chromosome in the population. As can be seen in Figure 3, once the best chromosome was found the frequency of the node of the best chromosome in the population kept increasing.

This indicated that chromosomes that included nodes from the best chromosome (largest clique) were more likely to be selected from the next generation.

In Figure 4, the best chromosome was found in the very last generation. This happens when the GA finds a clique of size that is equal to the best one known. In this case, the GA short-circuits and terminates. However, we observe that the frequency of the nodes in the (eventual) best chromosome increases through the generations. This indicates that, at each generation, the GA was selecting cliques with nodes from the winning chromosome.

## Conclusions and Final Remarks

Our results show that, in general, stochastic techniques may often outperform adaptive-search techniques such as genetic algorithms on the DIMACS datasets. The relatively simple Monte Carlo algorithms we developed often found larger cliques than the Zhang FGA and, very often, faster. However, it was not a clean sweep. For some classes of graphs, the genetic algorithm outperforms the Monte Carlo algorithms. This indicates that recombination sometimes works much better than purely random search. The graphs in the DIMACS dataset are varied. We noticed that the MC algorithm sometimes struggles when the graphs are very dense. We conclude that purely stochastic search should not be dismissed when approximating NP-Hard graph problems.

While genetic algorithms have the potential to be very powerful optimization tools, their success is highly dependent on the careful design and tuning of genetic operators, maintenance of population diversity, and accurate fitness evaluations. If the genetic operators do not preserve high-quality traits from parent solutions, the algorithm can fail to improve over generations. In contrast, stochastic search methods, due to their simplicity and robustness, can sometimes outperform genetic algorithms, particularly when the latter are not well-suited or finely tuned to the problem at hand. Stochastic search methods are generally easier to implement and do not require fine-tuning of complex genetic operators. Furthermore, stochastic methods typically do not suffer from premature convergence as they continuously explore new areas of the solution space. To maximize the effectiveness of genetic algorithms, it is crucial to address these common pitfalls through tailored genetic operators, adequate parameter tuning, and a large population size (to maintain diversity). For many optimization problems, this might not be straightforward.

## Use of Generative AI

Generative AI was not used in the preparation of this paper.

## Acknowledgements

Special thanks, and gratitude, to the Department of Computer Information Systems in the Faculty of ICT at the University of Malta for making available the computational resources required for this work.